# An Attention-Based Deep Learning Architecture for Real-Time Monocular Visual Odometry: Applications to GPS-free Drone Navigation

Olivier Brochu Dufour[1], Abolfazl Mohebbi[1*], Sofiane Achiche[1]

**Abstract**

   Drones are being used increasingly in many fields of application such as industry, medicine, research, disaster relief, defence, and security sectors. However, technical limitations like navigation in GPS-denied environments can be a barrier to further adoption of this technology. Research in visual odometry is rapidly evolving and could provide solutions to GPS-free drone navigation. Current visual odometry techniques use standard geometry-based pipelines. Although popular, existing solutions can suffer from significant drift caused by an accumulation of errors and can be computationally expensive. New research using deep neural networks has shown promising performance and could eventually offer a solution to the shortcomings of existing geometry-based techniques. Deep visual odometry techniques commonly combine convolutional neural networks (CNNs) and sequence modelling networks like recurrent neural networks (RNNs) to build an understanding of the scene and infer visual odometry from a given video sequence. This paper describes a novel real-time monocular visual odometry model for GPS-free drone navigation using a deep neural architecture, whose main innovation is the use of self-attention module. This deep learning model estimates, from consecutive video frames, the ego-motion of a camera rigidly attached to a vehicle's body. An inference utility captures the live video feed from a drone and uses deep learning model predictions on the video frames to assemble a complete trajectory estimation. The architecture includes a convolutional neural network to perform image feature extraction and a long short-term memory (LSTM) network combined with a multi-head attention module to model the sequential dependencies of the video. The model is trained using two common visual odometry datasets. The results indicate that the proposed model converges 48% faster than another RNN-based visual odometry model, which has previously shown promising results. In addition, a reduction of 22% in mean translational drift from the ground truth and an improvement of 12% in mean translational absolute trajectory error was also observed. In the end, the proposed model also showed more robustness to noisy input.

[1] Department of Mechanical Engineering, Polytechnique Montreal
[*] Corresponding Author: <u>abolfazl.mohebbi@polymtl.ca</u>
Address: 2500 Chem. de Polytechnique, Montréal, QC H3T 1J4 Canada.

# 1. Introduction

Unmanned aerial vehicles (UAVs) are found in many application domains, given their flexibility and usefulness. However, it can be challenging to develop industry solutions for small, resource-limited drones that need to operate in environments with no available GPS data or for drones that require exact positioning [1, 2]. In addition, under normal circumstances, GPS is crucial for precise and reliable odometry data. Current solutions to these challenges use geometry-based computer vision algorithms to calculate the movements of a camera rigidly attached to the body of the vehicle. This process is called visual odometry [3, 4].

New research attempts to tackle these issues using deep neural networks with surprising success. Rather than handcrafted features, deep learning models extract and encode complex features from the input data to regress the camera pose. Some networks have reached state-of-the-art performances on popular datasets, but none are specifically designed according to the computational requirements of onboard computers [5]. Currently, most of the best-performing visual odometry algorithms, neural or otherwise, are either resource-inefficient and computationally expensive or require extra sensors (inertial measurement unit, second camera, etc.). A small and computationally efficient single-camera solution to visual odometry is essentially needed to address the limitations of navigating and operating drones in GPS-denied environments [5].

The main objective of this research is to develop an intelligent system capable of estimating the visual odometry of a drone in real time while considering the current technological limitations of small onboard computers and the desire to improve the state of visual odometry research. Consequently, we aim to design a supervised deep neural network capable of producing accurate visual odometry estimations on live video streams. We hypothesized that a Multi-head Attention Module [6] could increase the accuracy of visual odometry estimations. New research from the natural language processing domain shows that neural networks using attention modules can outperform other types of networks in sequence modelling tasks [7]. Since visual odometry is fundamentally a sequence modelling problem, it is possible that a multi-head attention module can help increase the accuracy of visual odometry estimations.

The rest of this paper is organized as follows: Section 2 reviews some related works, and Section 3 describes the proposed model in detail. The results regarding the performance of our approach are presented in Section 4 and compared with an efficient model from the literature. Finally, we conclude the paper in Section 5.

# 2. Related Work

## 2.1. Geometry-Based Methods

Visual odometry estimation using geometry-based methods was the first category of techniques developed in the 1980s to compute a vehicle's 3D motion in a scene, also called ego-motion [8]. The geometry-based algorithms can be divided into feature-based and direct methods, which can use monocular or stereoscopic configurations.

### 2.1.1. Sparse feature-based methods

Feature-based methods typically consist of camera calibration, feature detection, feature matching, outlier rejection, motion estimation, scale estimation, and optimization [9]. Although popular, properly formulating and detecting complex features for recovering specific motions is still challenging. Therefore, research into visual odometry using sparse features methods has been heavily focused on improving the accuracy through various schemes such as error modelling [10], pose refinement [11] and outlier rejection [12], to name a few.

### 2.1.2. Direct methods

Unlike feature-based methods, direct methods skip the pipeline's feature extraction and matching steps entirely and instead process the whole image directly in the motion estimation step. Direct methods achieve this by minimizing an error metric that uses information from the pixels of consecutive images to obtain the camera's rigid body transformation [13]. For example, common methods like DTAM (Dense Tracking and Mapping) [14] and LSD-SLAM (Large-Scale Direct Simultaneous Localization and Mapping) [15] minimize the photometric error. Although there is a wide variety of error metrics, most rely on the brightness consistency constraint and a global motion model [16] to further constraint the overall motion estimate.

### 2.1.3. Performance of geometry-based methods

Both direct and feature-based visual odometry techniques are considered state of the art and choosing between them depends on the specific requirements of the problem at hand. Feature-based techniques are superior in highly textured scenes since it is easier to extract robust features. Furthermore, since these algorithms only extract a few hundred points from an image, their tracking step is usually fast. However, extracting those data points can be computationally expensive, depending on the scene and the algorithm. Due to their outlier rejection mechanism, feature-based methods are also very robust when dealing with dynamic elements in a scene, such as pedestrians. However, their performance significantly drops when confronted with low-texture scenes since it becomes difficult to extract reliable features from them [17].

Direct methods are more robust when presented with texture-less scenes because they ingest the complete image, unlike feature-based methods. Since they do not carry out feature extraction, description, and matching, most of the computing effort is spent tracking and mapping the scene - a process more computationally expensive than feature-based methods. Lastly, they are susceptible to illumination changes because these methods assume consistent lighting intensity, a factor that can be hard to control in real-world use cases.

Benchmark data from the KITTI visual odometry leaderboard show that stereo vision algorithms offer much greater accuracy than their monocular counterpart [18]. This is because stereo systems can have a better 3D understanding of the scene and, therefore, much more easily recover its scale, making them less susceptible to the accumulation of scale drift over time, leading to inaccuracy [3, 19, 20]. However, even though cameras are now more affordable, not all systems are equipped with stereoscopic vision systems and have the computational power to do so.

Therefore, improvements in monocular vision systems, which are more common and usually have lower computational costs, are still of interest to the scientific community.

### 2.2. Deep Learning-Based Methods

As discussed earlier, the performance of standard monocular odometry techniques is inferior to that of stereoscopic techniques. This is attributable to their misperception of the scene's three-dimensional geometry, which leads to scale errors and drift. However, there is still significant interest in overcoming these hurdles, and researchers have turned to deep neural networks as a potential solution to monocular odometry estimation. In the literature, supervised and unsupervised models are both used with varying degrees of success.

*2.2.1. Supervised models*

Early work by Konda et al. [21] investigated convolutional neural networks (CNN) for visual odometry by attempting to extract the change in velocity and direction from the estimated depth of a stereoscopic sequence. Unfortunately, it characterized the problem as classification instead of regression, limiting the network's performance. In later work, DeepVO by Wang et al. [22] further explored the use of CNN for visual odometry by coupling it with a long short-term memory (LSTM) network to extract sequential dependencies and infer pose from raw images. Apart from the novel addition of the LSTM network, DeepVO also introduces the use of FlowNet [23] convolutional architecture and weights for the convolutional layers. When trained, the convolutional layers learn to extract features similar to FlowNet's features. MagicVO by Jiao et al. [24] builds upon DeepVO architecture but uses a bidirectional LSTM to allow the network to learn information from the previous time steps and future ones. MagicVO and DeepVO are monocular networks and, as such, are affect by scale drift [3]. More recently, Fang et Al. [25] introduced two new regularizing losses: a Graph Loss and A Geodesic Rotation Loss, modestly improving performances of monocular visual odometry networks.

Another common strategy is to use auxiliary task learning to reinforce consistency between predicted outputs. For example, Valada et al. (VLocNet) [26] showed how the problem of monocular visual odometry could be formulated as an auxiliary task to global pose regression. The proposed architecture uses a CNN network to predict odometry, which shares parameters with a second network used to predict global pose. A loss that enforces geometric consistency between the outputs is used to jointly train both networks. The follow-up paper describing VLocNet++ by Radwan et al. [27] expands this idea of auxiliary learning by introducing a large neural network, achieving state-of-the-art performances, that is jointly optimized to predict visual odometry, global pose estimation and semantic segmentation. Interestingly, not only does the network output semantic segmentation, but it also uses it to further refine its pose estimation by focusing its attention on more informative regions of the scene.

Lin et al. [28] explored auxiliary task learning by jointly training two recurrent convolutional neural networks (RCNNs) to predict both global and relative pose, using a loss that enforces temporal geometric consistency between the two predictions. Parisotto et al. [29] proposed an architecture inspired by DeepVO but replaced the traditional RNN with temporal convolutions.

The estimated relative poses are assembled into global poses, which are then refined by a "neural graph optimizer" composed of a series of multi-head attention modules [30] and temporal convolutions.

*2.2.2. Unsupervised models*

The common motivations for using unsupervised learning are the lack of access to properly labelled datasets. Not only are unsupervised networks a great solution to this problem, but they also offer competitive performances compared with state-of-the-art methods [18]. SfMLearner by Zhou et al. [31] introduced the use of view synthesis as a supervisory signal to learn both depth and pose in an unsupervised fashion. The idea builds on an earlier paper [32] that demonstrates that synthetically generated points of view of a scene can be used as a metric to evaluate the quality of optical flow and stereo correspondence estimations. The network is forced to learn the proper camera motion to reconstruct the scene from a different position. Although it shows competitive results, as mentioned in the paper, the performances of the network can be strongly affected by moving objects, occlusions, and non-Lambertian surfaces [33].

UnDeepVO by Li et al. [34] introduced a similar network to SfMLearner with a different training pipeline. Notably, it trains on a stereoscopic image pair, using both images to compute the left-right photometric consistency loss and a left-right pose consistency loss. They also use the right images to calculate the photometric consistency loss of consecutive monocular images and the 3D geometric registration loss of consecutive monocular images. UnDeepVO is trained on stereoscopic images, and inferences are made on monocular images. UnDeepVO can approximate scale, which makes it less prone to scale drift and, thus, more accurate than purely monocular approaches [3].

Contrary to SfMLearner and UnDeepVO, Lyer et al. [35] introduced a lighter network that did not learn depth as an auxiliary task. Instead, it uses an RCNN architecture trained using the Composite Transformation Constraints loss. The loss forces the network to produce pose estimations that are geometrically consistent and follow the law of composition of rigid body transformations. Unfortunately, generalization to unseen data is poor. D3VO by Yang et al. [36] jointly learned to estimate depth, relative pose and the photometric uncertainty between input images. By predicting the photometric uncertainty, the network reduces the weights of the pixels that violate the brightness constancy assumption [37]. This novel proposal allows the network to outperform other unsupervised monocular deep visual odometry networks on the KITTI visual odometry leaderboard [18].

Although unsupervised deep visual odometry techniques currently outperform supervised techniques on the KITTI dataset, they result in larger networks that are not necessarily best suited for power- and memory-restricted vehicles like drones. We hypothesize that attention-based layers could improve the performance of deep monocular visual odometry networks without significantly increasing their size and complexity.

## 3. Methods

### 3.1. Problem Statement

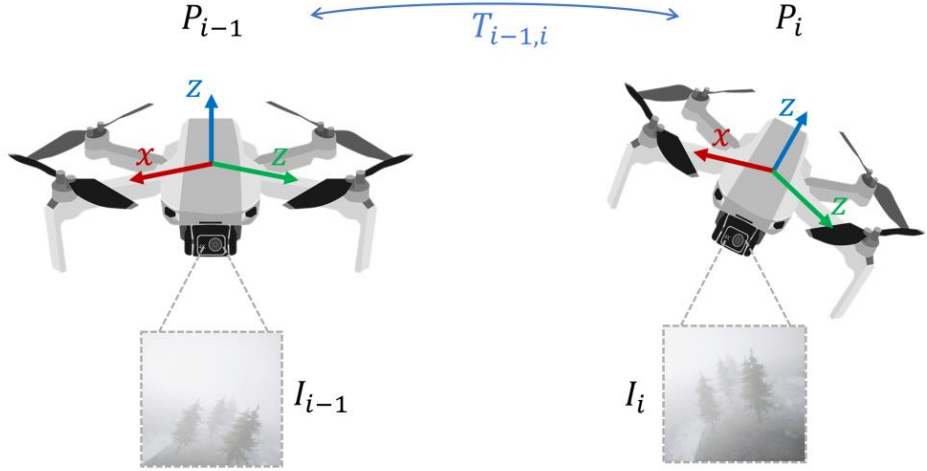

Figure 1 – The description of the transformation $T_{i-1,i}$ between vehicle poses $P_{i-1}$ and $P_i$. The transformation $T_{i-1,i}$ is a function of images $I_{i-1}$ and $I_i$ from a camera rigidly attached to the body of a drone.

Given a camera rigidly attached to the body of a drone and sharing the same coordinate frame, as shown in Figure 1, then the rigid body transformation of the camera from time step $i - 1$ to time step $i$ can be expressed as a function of image inputs $I_{i-1}$ and $I_i$:

$$T_{i-1,i} = f(I_{i-1}, I_i), \ T_{i-1,i} \in R^{4\times 4} \quad (1)$$

The transformation matrix $T_{i-1,i}$ can be composed with the previous transformation to obtain pose at time step $i$:

$$P_i = P_{i-1} T_{i-1,i} \quad (2)$$

where $P_i \in R^{4\times 4}$ is the pose consisting of the vehicle's position and orientation at time step $i$. Then the set of all poses from time step 0 to $n - 1$, is the complete trajectory of the vehicle computed from the sequence of images $I_0$ to $I_{n-1}$.

This project aims to build a neural network that can approximate $f(I_{i-1}, I_i)$ by learning the probability distribution $p(T_{i-1,i} \vee I_{i-1}, I_i)$ that best describes the training data.

### 3.2. Datasets

The KITTI [38] dataset and the Mid-Air [39] dataset were selected for this study as they offer a wide variety of scenes captured from an RGB camera rigidly attached to a moving vehicle. The datasets provide 6 Degrees of Freedom (DOF) ground truth pose data recorded by GPS/IMU sensors. Regarding the KITTI dataset, the video sequences from the left color camera were used.

The camera sensor runs at 10 Hz, and the final image resolution is $1328 \times 512$. To be usable, the ground truth pose data is projected in the left camera frame. The dataset was segmented so that sequences 00, 01, 02, 05, 08 and 09 are used as the training set, sequences 03, 04 and 06 as the validation set and sequences 07 and 10 as the test data.

As for the Mid-Air dataset, all the trajectories provided were used. The Mid-Air dataset offers 30 training trajectories in four different weather conditions (sunny, sunset, foggy, cloudy) and 24 training trajectories in three different seasons for a total of 216 training trajectories. Furthermore, it has five validation trajectories in four weather conditions and six validation trajectories in three different seasons for a total of 38 validation trajectories. Lastly, the dataset has three test trajectories in three climate conditions (foggy, sunny, sunset) for a total of nine test trajectories for the purpose of benchmarking. The images used in this project are provided by the left RGB camera, which sampled the scenes at 25 Hz and at a $1024 \times 1024$ pixel resolution. All sensors (GPS IMU, cameras) are aligned at the body frame's origin.

### 3.3. Training Environment

The model was built in Python 3.6.10 using PyTorch 1.6.0. It was trained on Compute Canada's Cedar cluster using an NVIDIA V100 GPU node running CUDA 10.1. Using a compute cluster allowed for an extensive hyperparameters search; however, the network is light enough to be trained and run on consumer-grade GPU, provided it has enough memory.

### 3.4. Training Procedure

The training procedure is broken down into three steps: *data acquisition and preparation*, *training*, and *evaluation*.

In the *Data Acquisition and Preparation* step, we decompose complete image sequences of trajectories into segments of 5 to 7 frames. Although it is desired for the final system to compute the odometry of a vehicle on complete trajectories, it is impractical to train the network on the raw trajectories themselves due to the memory limitation.

Since the ground truth data of each trajectory segment does not start at the origin, the world frame and camera frame are realigned to the origin at the start of the segment. During this step, we have also computed the dataset's mean and standard deviation for all three image channels. These are used to normalize the image's pixel values. The images are resized to $608 \times 184$ to reduce their memory footprint. Finally, the Mid-Air dataset's frame of reference is transformed into the KITTI dataset's camera frame for consistency.

During the *Training* step, data is acquired and fed to the model in batches that contain a randomly sampled collection of segments to allow the training process to use parallel processing on GPUs. We found that a mini-batch size of N=15 maximizes both memory usage and accuracy [40]. The model is trained to optimize the mean squared error (MSE) between the ground truth and the network's estimation using Adagrad [40] with a learning rate of 0.0005. The MSE between the estimation and the ground truth is separately evaluated on the rotation and translation components of the output vector. The error is estimated over the whole minibatch. The mean error of the rotation and translation estimates are summed, and the rotation estimate is scaled by 100 to increase its weight in the training signal. Equation (3) describes the procedure:

$$Loss_{MSE} = \frac{1}{N} \sum_{j=0}^{N-1} \left( \sum_{i=0}^{n-1} 100 * \|\hat{\phi}_i - \phi_i\|_2^2 + \|\hat{t}_i - t_i\|_2^2 \right)_j \tag{3}$$

where $\hat{\phi}_i$ and $\phi_i$ are the predicted and ground truth Tait-Bryan angles, respectively. Here, $i$ is the index of the timestep of the pose and $j$ is the index of the segment of the minibatch of size $N$.

Finally, in the *Validation* step, we evaluate the ability of the network to generalize to unseen data at the current stage of the training process using the validation set. The same loss function used in training is used as the evaluation metric for model validation.

### 3.5. Testing Procedure

Unlike the validation steps of the training procedure, the final model performances are evaluated on the complete test trajectories instead of trajectory segments. An overlapping sliding window mechanism is used to scan a trajectory and generate input sequences of predefined size for the model. The resulting predictions are then assembled into a continuous set of poses, using Equation 2. Picking the right size for the sliding window and the number of overlapping frames is a trade-off between the memory size, the inference accuracy of the model, and the desired speed of trajectory inference. Large sliding windows require larger memory but allow for more data to be used in a single inference pass. This is useful if the model's accuracy on short sequence is low because, theoretically, a neural network that can model long-term dependencies can adjust its prediction by observing more of the input sequence. A large frame overlap allows the model to consider more of the past context given new input frames, but it increases the inference time because more frames need to be recomputed. For this project, tests showed that a sliding window size of 30 and an overlap of 15 yielded a good accuracy vs. speed balance. Computed trajectory poses are evaluated using the KITTI Error (KE) metric introduced with the KITTI dataset and the Absolute Trajectory Error (ATE). Performance metrics are computed on both individual trajectories and the complete set of test trajectories. The KITTI error metrics are designed to give information about the amount of drift the model will accumulate over a set of discrete distances (100 m, 200 m, etc.) for both translation and rotation estimates. They are defined as follows:

$$KE_{rot}(P_d) = \frac{1}{d} \sum_{(i,j) \in P_d} \angle \left[ \left( \hat{P}_i^{-1} \hat{P}_j \right)^{-1} \left( P_i^{-1} P_j \right) \right], \tag{4}$$

$$KE_{trans}(P_d) = \frac{1}{d} \sum_{(i,j) \in P_d} \left\| \left( \hat{P}_i^{-1} \hat{P}_j \right)^{-1} \left( P_i^{-1} P_j \right) \right\|_2. \tag{5}$$

From above, $P_d$ is a set of pose pairs $\{(P_i, P_j), (\hat{P}_i, \hat{P}_j)\}$, where $i$ is the starting frame of a segment of distance $d$ and $j$ is the end frame of that segment. $P \in R^{4 \times 4}$ is the transformation matrix of the ground truth pose, and $\hat{P} \in R^{4 \times 4}$ is the transformation matrix of the estimated pose.

$\angle[.]$ is the angle given by the arccosine of the trace of the rotation matrix. $\|.\|_2$ is the $L^2$-norm of the translation vector.

The Absolute Trajectory Error (ATE) compares two aligned trajectories using the root mean square error (RMSE) and is defined as follows for rotation and translation:

$$ATE_{rot}(P) = \sqrt{\frac{1}{n}\sum_{i=0}^{n-1}\|\angle(\hat{R}_i R_i^{-1})\|^2} \tag{6}$$

$$ATE_{trans}(P) = \sqrt{\frac{1}{n}\sum_{i=0}^{n-1}\|t_i - \hat{t}_i\|^2} \tag{7}$$

where $P$ is the set of prediction ($\hat{P}_i \in R^{4\times 4}$) and ground truth ($P_i \in R^{4\times 4}$) poses making the complete trajectory. $\hat{R}_i \in SO(3)$ and $R_i \in SO(3)$ are the aligned estimated and ground truth rotation matrices, respectively. $\hat{t}_i \in R^{3\times 1}$ and $t_i \in R^{3\times 1}$ are the aligned estimated and ground truth translation vectors, respectively.

### 3.6. Regularization

Three regularization techniques are used to help the deep neural networks better generalize on unseen data: artificial dataset augmentation, early stopping and dropouts. For the dataset augmentations, random variations in image brightness, saturation, and contrast are applied. Moreover, holes of random sizes are cut out of the image at random locations. Both types of perturbations are randomly applied to the images. To prevent overfitting, the training session is pre-emptively stopped when no improvement in training loss is observed after 15 epochs. Finally, dropouts are used to prevent the network from strengthening neural connections too quickly and instead force it to explore alternative connections configuration during the training process. Dropout is a regularization technique that can be seen as introducing noise into the input signal. This is effectively done by randomly severing neural connections between neural layers.

### 3.7. Neural Architecture and Model Design

SelfAttentionVO is composed of four main modules: (1) the convolutional neural network, which extracts features relevant to visual odometry from two stacked consecutive images, (2) a bidirectional LSTM module which transforms the extracted features into a time-dependent vector representation of the camera's ego-motion, (3) an attention module that adjusts this vector by weighting its component according to the context of the sequences, (4) two fully connected linear layers that reduce the multi-dimensional vector into simple $6 \times 1$ vectors composed of the camera's rotation and translation. This innovative architecture builds upon the success of previous research by combining a convolutional neural network and an LSTM [22, 24], while its attention module is a novel contribution. Figure 1 gives an overview of the architecture.

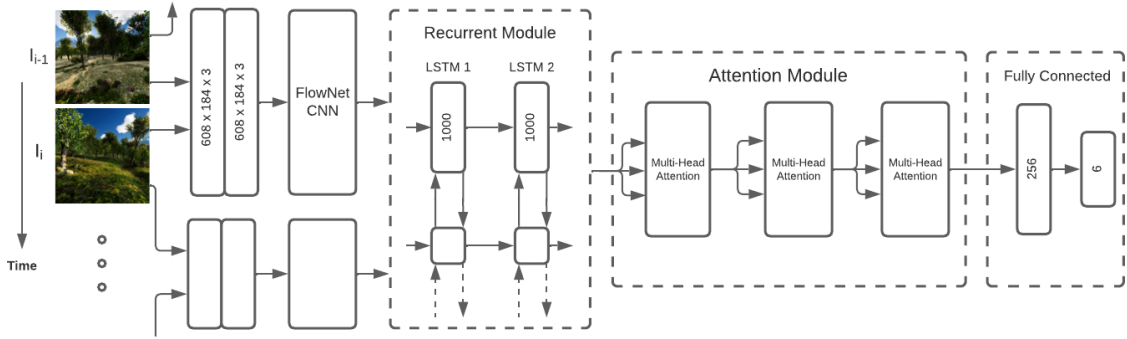

Figure 2 - An overview on the architecture of the proposed SelfAttentionVO

Attention-based architectures are becoming increasingly popular for signal processing tasks due to their efficiency. Yet they are under-researched in the field of visual odometry. According to the authors' best knowledge, only [29] currently implements an attention-based architecture for visual odometry tasks. Furthermore, coupling contextual pose vector representations generated by bidirectional RNNs with an attention module is original to this study.

Convolutional neural networks (CNNs) are a feedforward neural network that employs matrix convolutions, to ingest grid-like data such as images. They are useful to reduce and filter the input to identify patterns [41]. SelfAttentionVO uses the architecture and pre-trained weights of the convolutional encoder of FlowNet [23]. In FlowNet, these layers extract optical flow features from a stacked pair of images. The idea is that optical flow estimation and ego-motion estimation are related problems and that pixel motion can be used to estimate how a camera moves from one frame to the next. Each convolution is followed by a batch normalization, Leaky Rectified Linear Unit (LeakyReLU) activation [42] and a dropout layer.

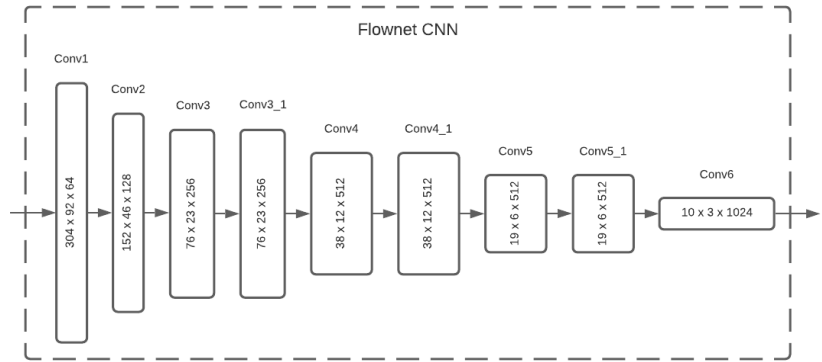

Figure 3 - An overview of the Flownet CNN

Recurrent Neural Networks (RNN) are used to model sequential data. They produce temporally dependent vector embeddings through a recurrent connection that feeds back into the network its previous hidden state. A Long Short-Term Memory (LSTM) network is a specialized RNN designed to model long term sequential dependencies [41]. Inspired by [24], SelfAttentionVO uses two stacked bidirectional LSTM networks. These LSTMs transform the extracted CNN features into a high-level vector representation of the camera's motion between time steps. Each vector is context-dependent and strongly influenced by the hidden state of adjacent LSTM cells. Each layer is composed of a recurrent bidirectional LSTM cell with a hidden state size of 1000 for each direction resulting in output vectors of size 2000. In addition, each layer output is subject to dropouts to aid in the network regularization.

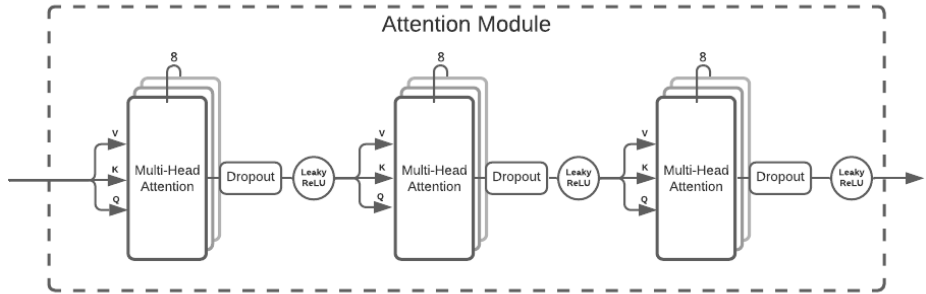

Figure 4 – Attention module, allowing the SelfAttentionVO model structure to refine the sequential dependencies modelled by the recurrent module

Attention networks can model very long sequential dependencies. Using a scaled dot product between the vector inputs, the network can compute the similarity between the vector representation of different timesteps from sequence and selectively ignore non-contextually relevant parts of an input vector [30]. The attention module, pictured in Figure 4, allows SelfAttentionVO to refine the sequential dependencies modelled by the recurrent module. The module takes as input the complete sequence of vectors outputted by the bi-LSTMs. Each multi-head attention layer will build a similarity matrix between the vectors. For SelfAttentionVO, the optimal configuration was found to be three multi-head attention layers composed of eight attention heads each. The attention heads will attend to information at different positions on the vector. Each multi-head attention layer re-enforces previously modelled sequential decencies. This creates pressure for the network to identify and track patterns in the sequence and tune out the noise in the signal. Dropouts are used both inside the attention heads and after to regularize the signal. LeakyReLU is used as the activation function after a layer.

Lastly, the fully connected layers are used by SelfAttentionVO to transform the high-dimensional pose transformation vector, tuned by the attention module, into a $6 \times 1$ vector describing the motion between frame $I_{i-1}$ and $I_i$. The first three elements of the output vector represent the rotation in the form of $y, x', z''$ intrinsic Tait-Bryan rotations and the last three elements represent the $x, y, z$ translation. The use of an intermediary layer that outputs a vector of 256 units allows for a more flexible compression of the attention module's output. This layer is followed by a dropout mask for regularization and *a LeakyReLU* activation function that introduce non-linearity in the signal.

## 4. Results

To facilitate the analysis of the results, an implementation of DeepVO [22] is used as a benchmark model against which the results obtained by the candidate models are compared using the same training environment.

### 4.1. Training Results

SelfAttentionVO successfully converged on all datasets and data augmentation combinations attempted (KITTI, Mid-Air, Augmented KITTI, Augmented Mid-Air, KITTI and Mid-Air, Augmented KITTI and Mid-Air), but the best solution was found after only 21 epochs of training on the augmented combination of KITTI and Mid-Air. Training with a large number of epochs (250), as reported in the DeepVO paper [30], did not yield optimal solutions. In fact, validation loss data showed clear signs of training data overfitting. Because of this, there are no obvious benefits in training for large numbers of epochs and early stopping proved to be a good strategy to prevent data overfitting.

In fact, we discovered that SelfAttentionVO consistently converges faster to a solution compared to its counterpart during training sessions on many dataset configurations. For example, on the KITTI dataset, solutions were found on average after only 44 epochs with SelfAttentionVO. This is 48% faster than DeepVO, which took on average 84 epochs to converge toward a solution. Figure 2a shows that the SelfAttentionVO model achieved lower validation losses on the augmented combination of KITTI and Mid-Air datasets compared with DeepVO. Moreover, Figure 2b demonstrates that the average validation MSE loss for multiple training sessions of SelfAttentionVO on the KITTI dataset is lower than DeepVO.

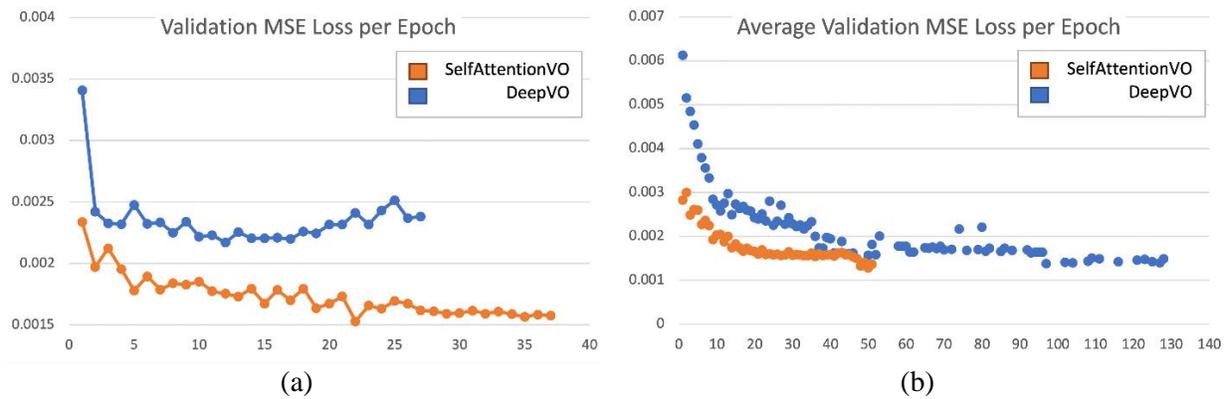

Figure 5 – (a) Validation MSE loss per epoch for the candidate (SelfAttentionVO) and benchmark (DeepVO) model - The candidate model and the benchmark model are trained on augmented data from the KITTI and Mid-Air datasets. (b) Average validation MSE loss per epoch (KITTI dataset) - The mean values are computed from the data of six training sessions for SelfAttentionVO and five training sessions for DeepVO on the KITTI dataset. The lowest average MSE loss for SelfAttentionVO is 0.00143 and 0.00172 for DeepVO.

## 4.2. Test Results

The optimal solution achieved during training for SelfAttentionVO yielded interesting performances on the test sets (non-augmented KITTI and Mid-Air test data). Results show that SelfAttentionVO performs well compared with the benchmark model DeepVO when trained and evaluated under the same conditions. The data reported in this section are calculated for test trajectories that have been truncated at 1,000 metres. Beyond 1,000 metres, the accumulated error becomes too great and skews the results. Only Mid-Air trajectories are affected by this measure since all the test trajectories of this dataset have lengths between 4,973 metres and 6,232 metres. SelfAttentionVO's performances can be observed in the KITTI translation and rotation error metrics, shown in Figure 3 and Table 1.

It is evident that the translational and rotational drift of SelfAttentionVO is lower than DeepVO, especially for segments less than 200m. This indicates that the local precision of SelfAttentionVO is higher than DeepVO. The impact of an estimation error is observed to be more significant on shorter path lengths. This is because a deviation of a few meters over a 100-200 m distance is proportionally more significant than over a few hundred meters. Since the amount of drift is relatively large for both models, error accumulation is substantial. However, the accumulation of errors does not grow faster than the length of the trajectory, which is why a downward trend can be observed in the drift percentages on both the translational and rotational error graphs.

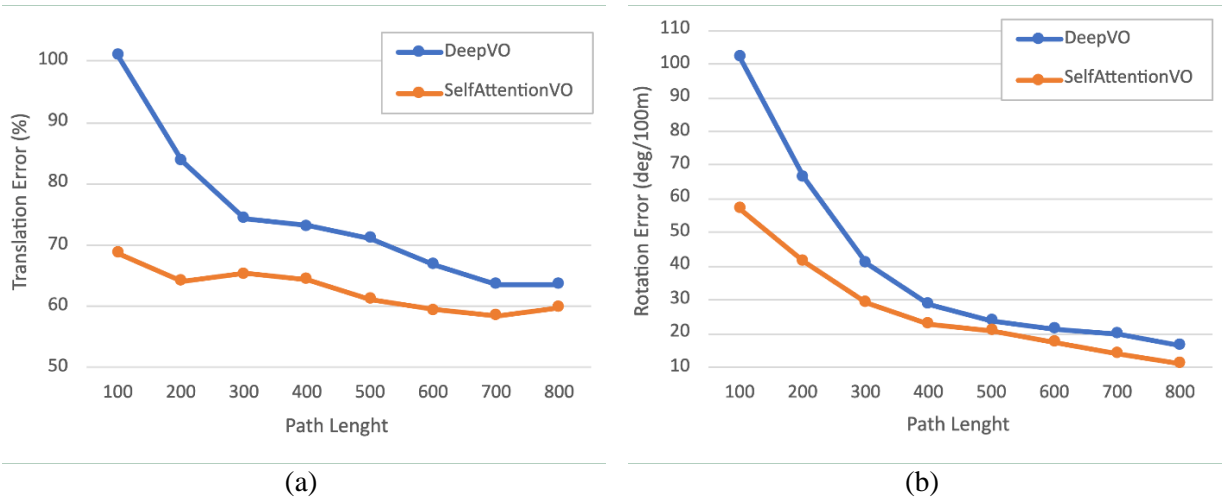

(a)                (b)

Figure 6 – Translation (a) and rotation (b) error per path length for DeepVO and SelfAttentionVO

Table 1 - Detailed average translation and rotation error per path length for DeepVO and SelfAttentionVO

|  | SelfAttentionVO | | DeepVO | |
|---|---|---|---|---|
| Path Length (m) | Translation Error (%) | Rotation Error (deg/100 m) | Translation Error (%) | Rotation Error (deg/100 m) |
| 100 | 68.6 | 56.9 | 101 | 102.1 |
| 200 | 64.1 | 41.2 | 83.8 | 66.3 |
| 300 | 65.4 | 29.3 | 74.4 | 40.7 |
| 400 | 64.4 | 22.6 | 73.1 | 28.4 |
| 500 | 61.3 | 20.6 | 71.1 | 23.9 |
| 600 | 59.4 | 17.3 | 66.9 | 21.3 |
| 700 | 58.4 | 13.8 | 63.5 | 19.9 |
| 800 | 59.8 | 11.0 | 63.5 | 16.5 |
| Mean | 62.7 | 26.6 | 74.7 | 39.9 |

The KITTI error metrics evaluate the average deviation from the ground truth on specific segment lengths. They do not assess how well a predicted trajectory fits the ground truth. To better illustrate the performance differences between the candidate model SelfAttentionVO and the benchmark model DeepVO, Figures 4 and 5 describe the resulting trajectories using the KITTI and Mid-Air test sets. When doing a qualitative characterization of the trajectories, we can observe that SelfAttentionVO generates trajectory estimations that have a better fit than DeepVO's benchmark trajectory estimations. The results are most eloquent on the KITTI test trajectories (07, 10).

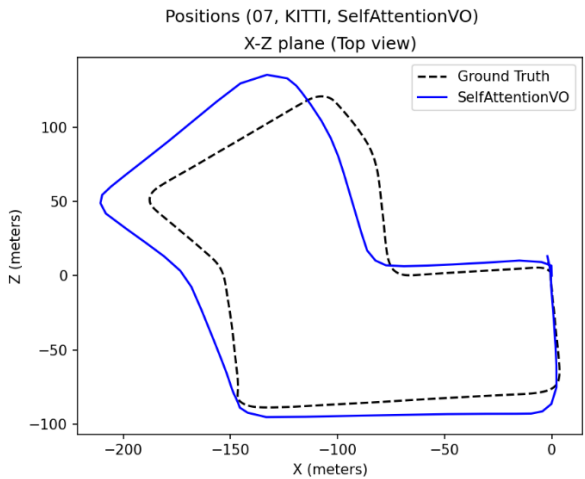

(a)

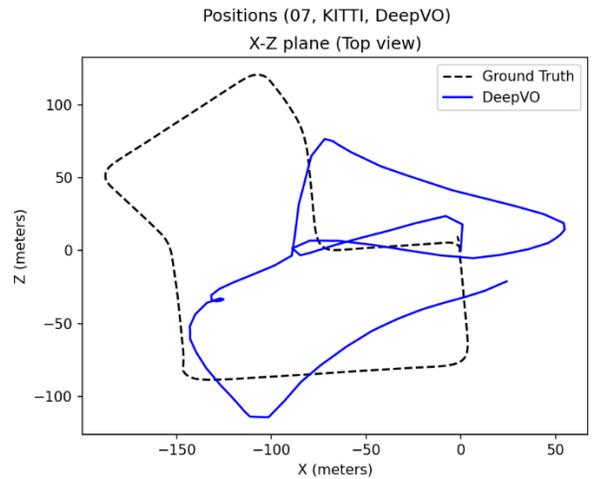

(b)

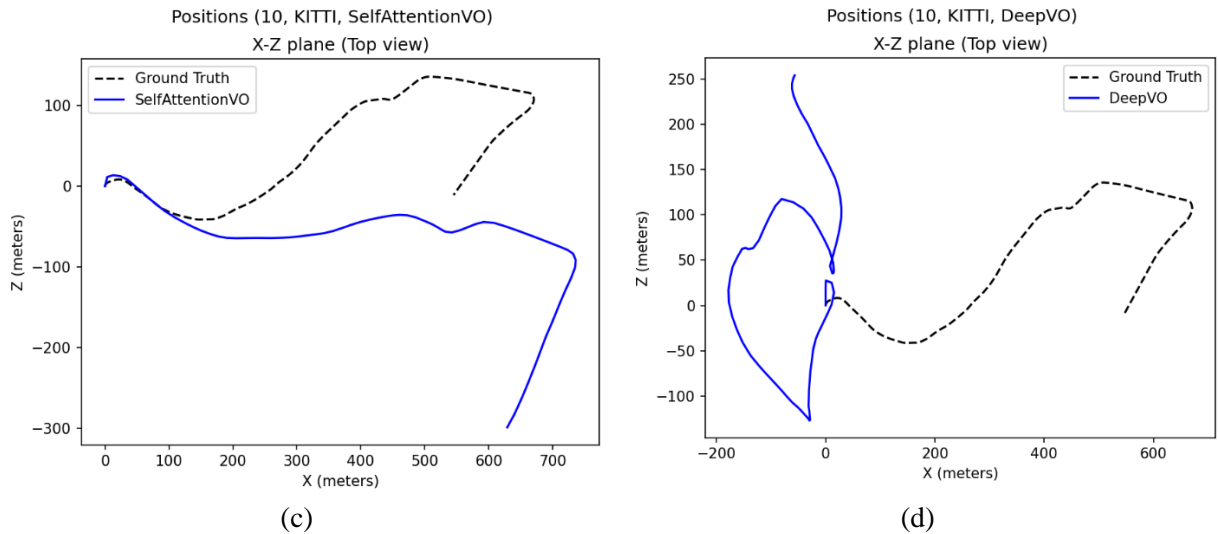

(c)                           (d)

Figure 7 –Translation Trajectory estimations by SelfAttentionVO and DeepVO for KITTI trajectories 07 and 10. The trajectories are viewed from the top, with the ground truth in dashed black and the estimated trajectory in blue.

Figure 6 shows a vertical view of KITTI 07 trajectory reconstructed by both models. It can be observed that, for both SelfAttentionVO and the benchmark model, the estimations suffer from significant error accumulation. It is unclear why this is the case, but we hypothesize that it could be caused by the lack of vertical motion in the KITTI dataset coupled with the low parallax of the large outdoor nature scenes of Mid-Air.

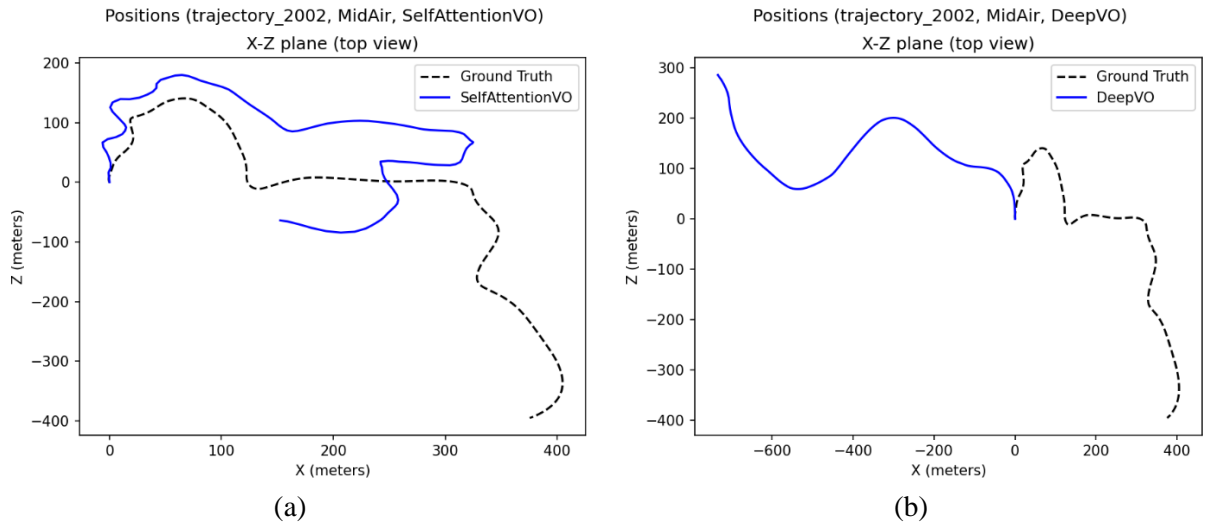

(a)                           (b)

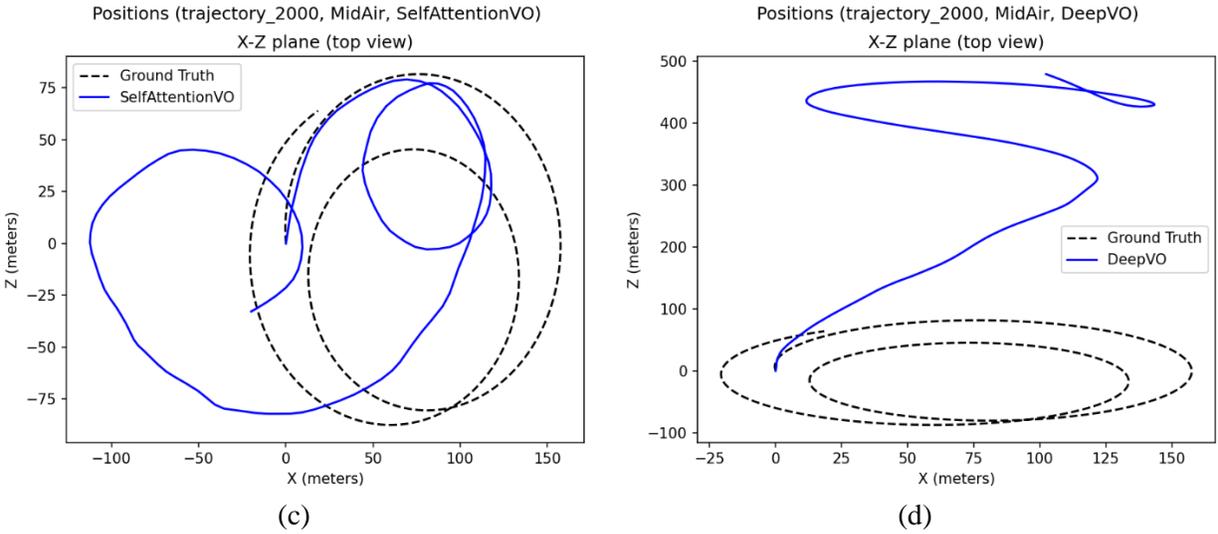

(c)                            (d)

Figure 8 – Trajectory estimations by SelfAttentionVO and DeepVO for Mid-Air trajectories 2000 and 2002. The trajectories are viewed from the top, with the ground truth in dashed black and the estimated trajectory in blue.

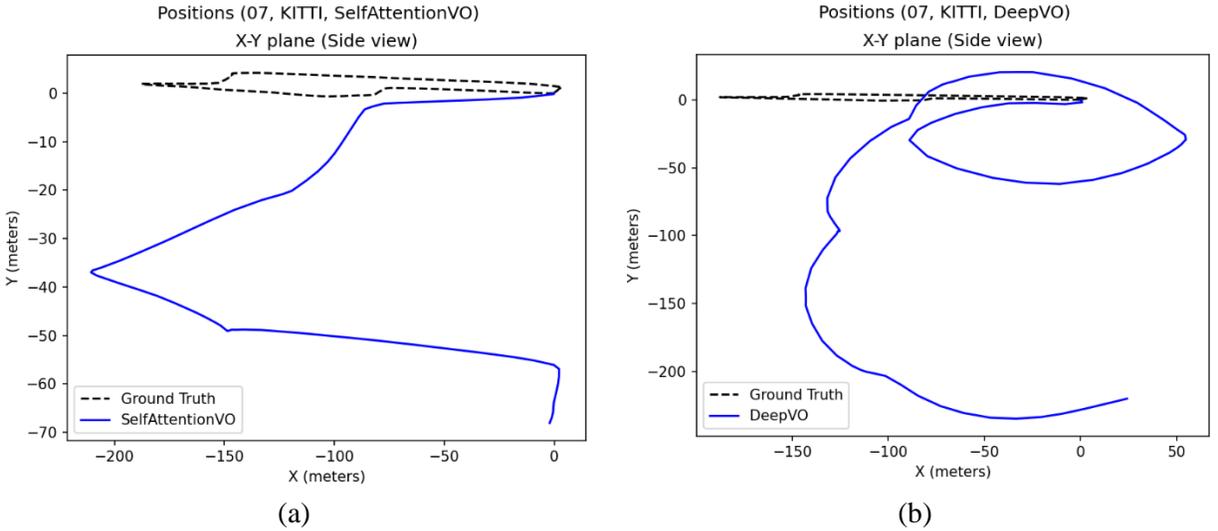

(a)                            (b)

Figure 9 – Trajectory estimations by SelfAttentionVO for KITTI trajectory 07 (vertical view). The trajectories are viewed from the side, with the ground truth in dashed black and the estimated trajectory in blue.

Table 2 shows the Absolute Trajectory Errors per trajectory. We can observe that SelfAttentionVO outperforms the benchmark model for most trajectories. There are significant improvements in SelfAttentionVO's ATE compared with the benchmark model on the KITTI dataset.

Table 2 - Mean Absolute Trajectory Error per test trajectory

| Trajectory | SelfAttentionVO | | DeepVO | |
|---|---|---|---|---|
| | Mean ATE Translation (m) | Mean ATE Rotation (deg) | Mean ATE Translation (m) | Mean ATE Rotation (deg) |
| KITTI 07 | 16.2 | 16.4 | 43.7 | 87.9 |
| KITTI 10 | 27.1 | 21.7 | 120.1 | 118.4 |
| Mid-Air 0000 | 62.8 | 87.2 | 66.5 | 113.4 |
| Mid-Air 0001 | 69.2 | 136.7 | 67.2 | 118.6 |
| Mid-Air 0002 | 103.2 | 60.4 | 68.2 | 91.3 |
| Mid-Air 1000 | 60.3 | 124.6 | 68.3 | 123.5 |
| Mid-Air 1001 | 65.5 | 126.5 | 67.0 | 131.6 |
| Mid-Air 1002 | 48.8 | 86.2 | 65.0 | 113.6 |
| Mid-Air 2000 | 56.4 | 58.2 | 68.3 | 125.6 |
| Mid-Air 2001 | 70.9 | 117.7 | 65.5 | 130.6 |
| Mid-Air 2002 | 89.0 | 51.0 | 60.0 | 111.3 |
| Mean | 60.8 | 80.6 | 69.1 | 115.1 |

To summarize, when trained on a combination of augmented KITTI and Mid-Air data, SelfAttentionVO's performances are better than the benchmark model. Overall, SelfAttentionVO allows for around 22% reduction in mean translational drift (KITTI Translation Error) and 40% reduction in mean rotational drift (Rotation Error) when calculated on complete trajectories (capped at 1,000 metres). Moreover, the translational fit is improved by about 12% (translation ATE) and the rotational fit is improved by about 30% (rotation ATE).

As discussed earlier, the attention module helps the network track patterns in the input sequence and tune out noise. The benefits of such an approach are quite eloquently illustrated in Table 3 which shows the performance of DeepVO and SelfAttentionVO when trained on non-augmented data but evaluated on augmented data. Since the type of data augmentation used is akin to data corruption, this table evaluates the performances of both networks on corrupted data. SelfAttentionVO appears to be significantly more robust to noisy/corrupted test data than DeepVO.

Table 3 - Model performance on corrupted data

| Model | Dataset | Augmented Test Data | Avg KITTI Translation Error (%) | Avg KITTI Rotation Error (deg/100 m) | Avg Translation ATE (m) | Avg KITTI Rotation ATE (deg) |
|---|---|---|---|---|---|---|
| SelfAttentionVO | KITTI | True | 61.2 | 53.7 | 73.8 | 121.9 |
| DeepVO | KITTI | True | 96.3 | 55.4 | 77.9 | 126.9 |

### 4.3. Real-Time Inference Performances

The visual odometry of a drone is particularly useful when computed in real time. Odometry information can be used in visual servoing systems and to inform decisions in tasks like autonomous drone guidance. To evaluate the real-time odometry estimation abilities of SelfAttentionVO, a real-time inference utility was designed. The utility provides an interface through which a real-time video feed is supplied and a real-time stream of odometry estimation is outputted. This utility was tested on a computer with an Intel Core i9-9900K CPU running at 3.60 GHz, and an Nvidia RTX 2080 Ti GPU.

To be usable, the solution should perform pose inference at a frame rate high enough to not drop any frames from the input feed. Using a sliding window of size 30 and overlap of 15, tests on a pre-recorded video (KITTI 07) streamed over HLS and UDP showed that, on average, the inference utility can process approximately 15 frames per second on the CPU, which is enough to process videos from the KITTI dataset. When running the utility on the GPU, the frame rate jumps to 60 frames per second, which is more than enough to process videos from the Mid-Air dataset. Since the sliding window configurations for live inference are the same as those used for the model testing, the estimation accuracy using the inference utility is technically the same as the accuracy observed during the testing phase. These tests show that the inference utility is a viable way of performing visual odometry estimation on a live video feed.

### 5. Discussion

The presented results demonstrate that an attention-based deep neural network like SelfAttentionVO can offer an interesting solution to real-time monocular odometry estimation for drones.

Although the results are very promising, significant shortcomings and issues still need to be addressed. Scenes with low parallax and points of interest located far from the camera are challenging for visual odometry algorithms since less data is available for ego-motion estimation. It can be observed from the results of Figure 4 and 5, that there is significant discrepancy between the quality of the odometry estimations on the KITTI dataset versus the Mid-Air dataset. The KITTI dataset includes videos taken from a car driving around a city for which the points of interest are relatively close to the camera, the car's movement is predictable and there is limited motion in the vertical axis. On the contrary, Mid-Air's videos are captured by a camera mounted on a drone, the motion of the drone is much less predictable and can be significant in all six DOFs. Moreover, the aerial footage means that most points of interest are distant. Overall, Mid-Air is a significantly more challenging dataset compared with KITTI due to its higher complexity.

Moreover, trajectory reconstruction appears to be most effective in the horizontal (x-z) plane and any reconstruction in the vertical axis (y) is affected by significant estimation errors for both SelfAttentionVO and DeepVO (Figure 6). One possible explanation is that the lack of vertical motion in the KITTI dataset coupled with the low parallax of the large outdoor nature scenes of Mid-Air. Further research is required to validate or invalidate this hypothesis. However, both trained DeepVO and SelfAttentionVO underperform during testing compared to DeepVO's published results. It is unclear what the source of this discrepancy is. However, similar discrepancies can also be observed in [25] which also attempted to reproduce DeepVO's results. Differences in hyperparameter selection may be to blame since DeepVO, like most other papers, does not publish its complete hyperparameter configuration.

Regardless of the shortcoming of SelfAttentionVO, the contribution of the multi-head attention modules cannot be overstated. Attempts to rely solely on the RNN for sequential modelling increased drift by about 50% and reduced fit by about 10%

## 6. Conclusion

The goal to design an intelligent system that can estimate the visual odometry of a drone in real time was successfully reached.

The main contribution to the field of visual odometry research is the novel architecture of the deep neural network. The neural network combines a convolutional neural network, a recurrent neural network, an attention module, and fully connected layers to extract the camera's ego-motion from a sequence of images. The convolutional network extracts visual features relevant to visual odometry, the recurrent network vectorizes those features into time-dependent vectors, the attention module tunes those vectors according to the context of the sequence and, finally, the fully connected layers compress those vectors into ego-motion predictions. The combination of the attention module and recurrent network to model the sequential dependencies is, to the best knowledge of the author, original to this research.

The network was trained on the KITTI and the Mid-Air dataset using a mean squared error loss. Test results showed that the architecture significantly reduces the mean translational drift and improves the mean translational fit of the estimated trajectory compared with DeepVO, a neural network often used as a benchmark in the literature for visual odometry. Even if the performances of this new model were equivalent to the benchmark model, the improvement in training time is a significant achievement and would certainly warrant more research on the use of attention mechanisms for visual odometry.

There are certain limitations to the results obtained by SelfAttentionVO. Namely, the network will require further tuning to bring its performance to the levels of accuracy reported in the literature. Moreover, testing on non-simulated aerial data is needed to characterize the network's performance on real drone footage. Strategies to improve the network's accuracy should be centred around loop closing and auxiliary task optimization. While the network can already be used through a real-time inference utility, its deployment onto an embedded onboard system is difficult. Future work should focus on optimizing the network's run-time performances so it can run in resource-limited environments.

Although the solution is designed to be used with drones, there is technically no specific architectural limitation that would prevent the proposed model from working with other types of vehicles, provided it is trained and tuned on the appropriate data. Given a properly trained model, the inference utility can be used with any kind of video input from various autonomous systems requiring visual odometry.

**Acknowledgments**

The project is funded by Natural Sciences and Engineering Research Council of Canada Grant Number RGPIN-2019-05497.